\documentclass[lettersize,journal]{IEEEtran}
\usepackage{amsmath,amsfonts}
\usepackage{algorithmic}
\usepackage{algorithm}
\usepackage{array}
\usepackage[caption=false,font=normalsize,labelfont=sf,textfont=sf]{subfig}
\usepackage{textcomp}
\usepackage{stfloats}
\usepackage{url}
\usepackage{verbatim}
\usepackage{graphicx}
\usepackage{epstopdf}
\usepackage{cite}
\usepackage{booktabs}
\usepackage{adjustbox}
\usepackage{multirow}
\usepackage{hyperref}
\hypersetup{hidelinks}
\usepackage{amsthm,amssymb}
\usepackage{mathrsfs}

\begin{document}
    \title{Trusted Video Inpainting Localization via Deep Attentive Noise Learning}

\author{Zijie Lou,
  Gang Cao,~\IEEEmembership{Member,~IEEE,}
  and~Man Lin
  \thanks{Zijie Lou, Gang Cao and Man Lin are with the State Key Laboratory of Media Convergence and Communication, Communication University of China, Beijing 100024, China, and also with the School of Computer and Cyber Sciences, Communication University of China, Beijing 100024, China (e-mail: \{louzijie2022, gangcao, lyan924\}@cuc.edu.cn).}}

\maketitle

\begin{abstract}
Digital video inpainting techniques have been substantially improved with deep learning in recent years. Deep inpainting can fill missing regions with plausible realistic contents. Although the inpainting is originally designed to repair damaged areas, it can also be used as malicious manipulation to remove important objects for creating false scenes and facts. As such it is significant to identify inpainted regions blindly. In this paper, we present a Trusted Video Inpainting Localization network (TruVIL) with excellent robustness and generalization ability. Observing that high-frequency noise can effectively unveil the inpainted regions, we design deep attentive noise learning in multiple stages to capture the inpainting traces. Firstly, a multi-scale noise extraction module based on 3D High Pass (HP3D) layers is used to create the noise modality from input RGB frames. Then the correlation between such two complementary modalities are explored by a cross-modality attentive fusion module to facilitate mutual feature learning. Lastly, spatial details are selectively enhanced by an attentive noise decoding module to boost the localization performance of the network. To prepare enough training samples, we also build a frame-level video object segmentation dataset of 2500 videos with pixel-level annotation for all frames. Extensive experimental results validate the superiority of TruVIL compared with the state-of-the-art. In particular, both quantitative and qualitative evaluations on various inpainted videos verify the remarkable robustness and generalization ability of our proposed TruVIL. Code and dataset will be available at \href{https://github.com/multimediaFor/TruVIL}{https://github.com/multimediaFor/TruVIL}.
\end{abstract}


\section{Introduction}
\IEEEPARstart{V}{ideo} inpainting aims to repair missing or damaged regions with plausible and coherent contents in a digital video. It has great value in many practical applications, such as scratch restoration\cite{chang2019free} and autonomous driving\cite{liao2020dvi}. However, video inpainting techniques may also be used to create forged videos by deleting or altering some contents, such as falsifying forensic evidences, removing copyright marks or key objects in news videos. Such malicious use of video inpainting potentially incurs societal risks and legal concerns. Moreover, with the impressive development of deep learning (DL), inpainted videos become more and more difficult to be distinguished, even to the naked eye. As shown in Fig. 1, key objects can be easily removed by the latest DL-based inpainting algorithms. Therefore, it is necessary to develop reliable forensic methods for identifying the inpainted regions in a video. 

In recent years, DL-based video inpainting approaches\cite{wang2019video, kim2020vipami, gao2020flow,li2022towards,zhang2022inertia, lee2019copy, oh2019onion, zeng2020learning, liu2021fuseformer, zhang2022flow} have made significant advancements. Through optical flow and attention mechanisms, the state-of-the-art methods can extract valuable temporal information from adjacent frames. This enables the created contents with high realism and spatiotemporal consistency. Nevertheless, such inpainting methods tend to acquire corresponding pixels from analogous regions or frames, or learn related distributions from similar scenes. As a result, there unavoidably leave behind some forensic traces and artifacts, such as regional inconsistencies, abrupt transitions  around object boundaries, and blurred areas caused by incomplete distribution acquisition. 

\begin{figure}[!t]
\centering
\includegraphics[width=\linewidth]{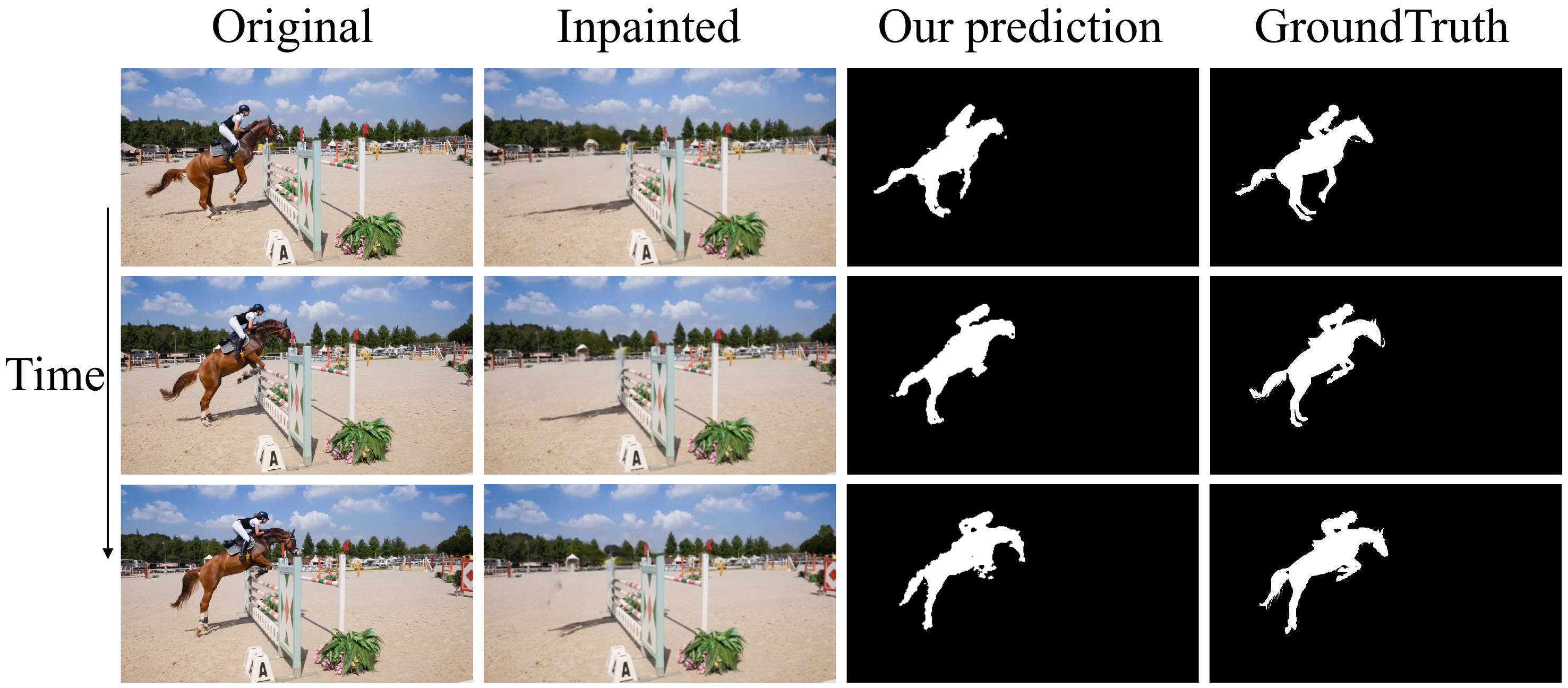}
\caption{Video inpainting localization. Given an inpainted video (second column), the inpainted regions are identified both spatially and temporally.}
\label{fig0}
\end{figure}

Some targeted forensic methods\cite{li2019localization, li2017localization, wu2021iid} have been proposed to fight against the malicious usage of inpainting manipulations, but they mainly focus on images, instead of videos. In addition, many universal forensic methods\cite{wu2019mantra, zhou2020generate ,dong2022mvss, wu2022robust, guillaro2023trufor, wu2023rethinking} have been developed for simultaneously detecting multiple types of image forgeries including splicing, copy-move, inpainting, etc. Although such image forensic methods could be applied to video inpainting detection in a frame-by-frame manner, they typically behave poorly due to the unexploited inter-frame correlation. Nowadays, several research papers on video forensics have been published\cite{zhou2021deep, yu2021frequency, wei2022deep, nguyen2022videofact, chen2015automatic, aloraini2020sequential, d2018patchmatch, feng2016motion}. The papers most relevant to our work are\cite{zhou2021deep, yu2021frequency, wei2022deep, nguyen2022videofact}, which share our focus on deep video inpainting localization. Zhou \textit{et al.} \cite{zhou2021deep} first present a LSTM-based video inpainting detection method, which combines RGB and error level analysis (ELA) information. The temporally consistent prediction is achieved by convolutional LSTM. Furthermore, Yu \textit{et al.} \cite{yu2021frequency} propose a spatiotemporal transformer framework to detect the spatial connections between patches and the temporal dependency between frames. The frequency domain information is exploited synchronously via a customized decoder. Moreover, Wei \textit{et al.} \cite{wei2022deep} extract the intra- and inter-frame residuals to reveal the spatial and temporal traces left by inpainting. The extraction of inter-frame residual is guided by optical flow for better exposure of inpainting traces. Nguyen \textit{et al.}\cite{nguyen2022videofact} propose a universal video forensic network - Videofact, which is able to detect a wide variety of video  forgeries. However, ideal performance is not achieved due to the complete ignorance of the temporal information among frames. Overall, such prior methods still leave a gap on forensic accuracy and reliability towards real-world applications. The robustness against post video compression and the generalization ability to unseen inpainting algorithms and datasets, though important in practice, have not been investigated in depth.

To attenuate the deficiencies of prior works, in this paper we propose a trusted video inpainting localization network (TruVIL) and perform extensive evaluations. We discover that the inpainted video regions can be exposed by high-frequency noise. Three novel modules are carefully devised to make full use of noise features. TruVIL is a two-stream network with two-modality inputs including RGB frames and noise features. To capture more rich and informative forensic traces, the multi-scale noise extraction module with HP3D layers is applied to the low-level RGB features at multiple scales. Then a cross-modality attentive fusion module is adopted to facilitate the interaction between two modalities. It benefits to share and exchange complementary forensic information, and enhance the representation learning of localization-towards features. Finally, an attentive noise decoding module is deployed to concentrate on more suspicious regions, thereby guiding the decoder properly and improving the final localization performance.

\begin{table*}[!t]
\caption{A taxonomy of the state-of-the-art deep learning-based image/video tampering localization methods. It indicates the used feature modalities and their fusion method, noise scales and backbone network. '-' denotes not applicable.}
\tabcolsep=1.3 pt
\label{tab0}
\begin{adjustbox}{width=\textwidth}
\begin{tabular}{rcllcl}
\toprule[1pt]
\multirow{2}{*}{\textbf{Methods}}         & \multicolumn{3}{c}{\textbf{Modalities}}              & \multicolumn{1}{c}{\multirow{2}{*}{\textbf{Scales of Noise}}}  &   \multicolumn{1}{c}{\multirow{2}{*}{\textbf{Backbone}}}  \\  \cmidrule(lr){2-4} 
                                            & RGB            & \multicolumn{1}{c}{Noise}         & \multicolumn{1}{c}{Fusion}           &          &     \\ \midrule
HP-FCN \cite{li2019localization}             & -              & High-pass filters                 &-                                     & -        & FCN\\
IID-Net \cite{wu2021iid}                     & \checkmark     & High-pass filters, BayarConv      &early stage (concatenation)           & single   & Adjustable CNN\\
CR-CNN \cite{yang2020constrained}            & -              & BayarConv                         &-                                     & -        & Mask R-CNN\\
GSR-Net \cite{zhou2020generate}              & \checkmark     &  -                                &-                                     & -        & Deeplabv2 \\
RGB-N \cite{zhou2018learning}                & \checkmark     & SRM filters                      &late stage (bilinear pooling)         & single   & Faster R-CNN \\
Mantra-Net \cite{wu2019mantra}               & \checkmark     & SRM filters, BayarConv            &early stage (concatenation)           & single   & Wider VGG\\
MVSS-Net \cite{dong2022mvss}                 & \checkmark     & BayarConv                         &late stage (dual attention)           & single   & FCN\\
TruFor \cite{guillaro2023trufor}             & \checkmark     & noiseprint++                      &middle stage (cross-modal calibration)& single   & Segformer\\
IF-OSN \cite{wu2022robust}                   & \checkmark     &  -                                &-                                     & -        & UNet\\
FOCAL \cite{wu2023rethinking}                & \checkmark     &  -                                &-                                     & -        & ViT, HRNet\\
EMT-Net \cite{lin2023image}                  & \checkmark     &  -                                &-                                     & -        & Swin-ViT\\
VIDNet \cite{zhou2021deep}                   & \checkmark     & ELA                               &early stage (concatenation)           & single   & VGG\\
FAST \cite{yu2021frequency}                  & \checkmark     & DCT                               &middle stage (concatenation)          & single   & ViT\\
STTNet \cite{wei2022deep}                      & \checkmark     & High-pass filters                 &middle stage (concatenation)          & single   & ResNet \\
VideoFact \cite{nguyen2022videofact}         & \checkmark     & -                                 &-                                     & -        & ViT\\
VIFST \cite{pei2023vifst}                   & \checkmark     & DCT                                 &middle stage (concatenation)        & single     & Transformer\\
UVL \cite{pei2023uvl}                   & \checkmark     & DCT                                 &middle stage (concatenation)        & single     & Transformer\\
CDSNet \cite{yao2024deep}                    & \checkmark     & SRM filters                       &middle stage (Add)                   & single    & ConvNeXt\\
TruVIL (ours)                               & \checkmark     & SRM filters                       &multi-stage                           & multiple & Uniformer\\
\bottomrule[1pt]
\end{tabular}
\end{adjustbox}
\end{table*}

Our major contributions are summarized as follows:

\begin{itemize}
\item{We propose a novel end-to-end network TruVIL for video inpainting localization, where inpainting artifacts are uncovered across spatial and temporal domains via deep attentive noise learning.} 
\item{We discover that the inpainted video regions can be exposed by high-frequency noise. To make full use of such feature, three functional modules including the multi-scale noise extraction, the cross-modality attentive fusion and the attentive noise decoding are devised elaborately.}
\item{TruVIL achieves much better localization performance in comparison with the state-of-the-art methods. Rigorous evaluations show its remarkable robustness and generalization ability.}
\item{A large frame-level annotated video object segmentation dataset is created for generating the inpainted video samples with normal frame rate. Such a dataset will be public to facilitate related experimental researches on the strict frame-level video object segmentation, editing and analysis, etc.}
\end{itemize}


\section{Related Works}
\subsection{Deep Video Inpainting}
Up to now, DL-based video inpainting algorithms\cite{wang2019video, kim2020vipami, gao2020flow,li2022towards,zhang2022inertia, lee2019copy, oh2019onion, liu2021fuseformer, zeng2020learning, zhang2022flow} have achieved impressive results in terms of both inpainting quality and speed. These methods can be divided into three categories: 3D Convolutional Neural Network (CNN)\cite{ kim2020vipami,wang2019video}, flow-guided\cite{li2022towards, gao2020flow, zhang2022inertia}, and attention-based\cite{oh2019onion, lee2019copy, liu2021fuseformer, zeng2020learning, zhang2022flow}. Wang \textit{et al.}\cite{wang2019video} propose the first DL-based video inpainting network, which comprises a 3D CNN for temporal prediction and a 2D CNN for restoring spatial details. However, 3D CNN is not adopted widely due to its high computational complexity. To address such concern, video inpainting is reconsidered as a pixel propagation problem which is solved by optical flow methodology\cite{li2022towards, gao2020flow, zhang2022inertia}. While such methods demonstrate promising results, they fall short in capturing the visible contents of long-distance frames. It leads to performance decline for dealing with the large or slowly moving objects. Correspondingly, some recent approaches\cite{oh2019onion, lee2019copy, liu2021fuseformer, zeng2020learning, zhang2022flow} adopt attention mechanism to capture such long-term contextual information. For example, Zeng \textit{et al.}\cite{zeng2020learning} propose to learn a multi-layer multihead transformer for video inpainting. Such a method is further improved by enhancing the boundary details of missing regions via soft split and composition\cite{liu2021fuseformer}. The rapid development of these deep video inpainting algorithms has significantly lowered the barrier for video editing. On the one hand, individuals can conveniently utilize deep video inpainting techniques to fill missing parts in videos. On the other hand, malicious attackers may exploit these techniques to manipulate videos for disseminating false information. Therefore, it is imperative to develop reliable video inpainting forensics algorithms to cope with the ever-evolving landscape of video inpainting algorithms.

\begin{figure*}[htbp]
\centering
\includegraphics[width=\textwidth]{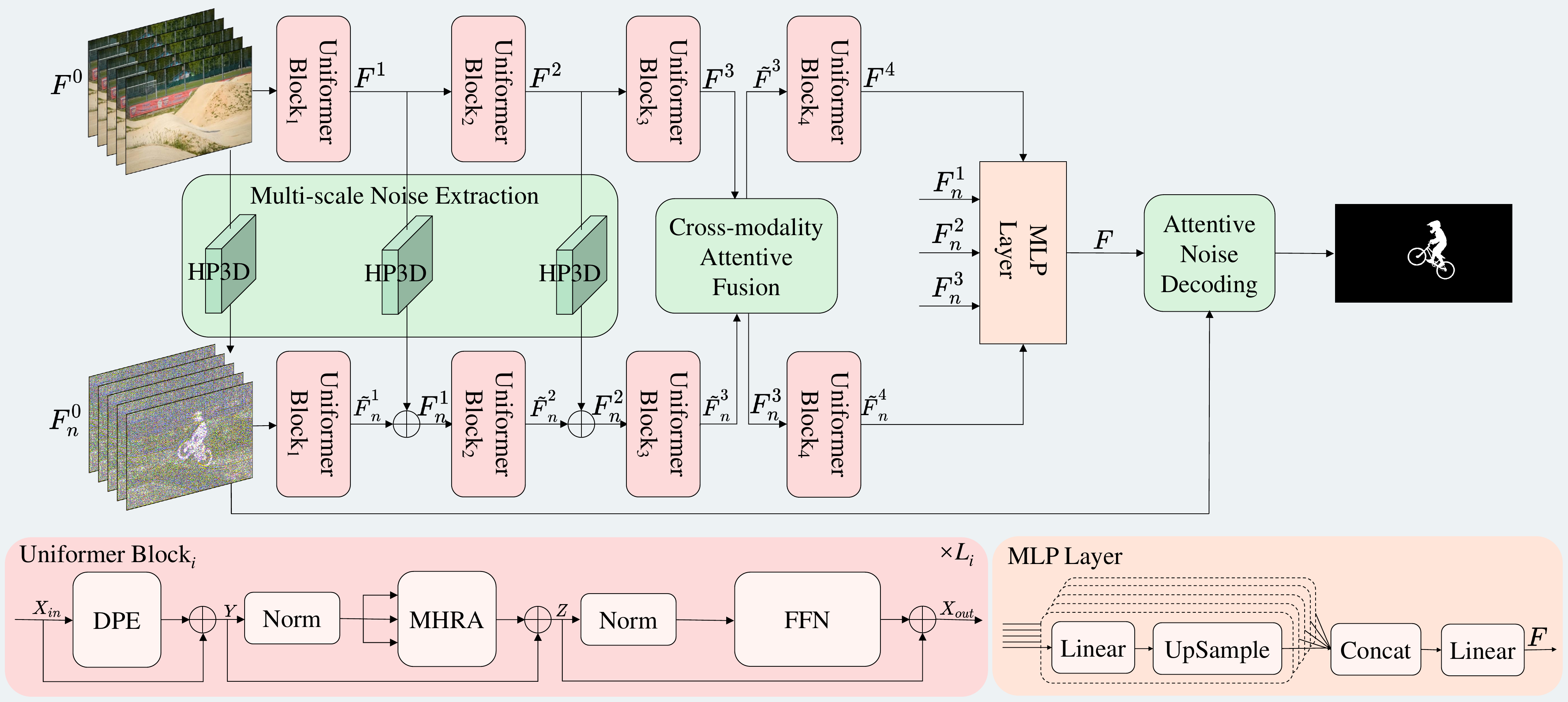}
\caption{Overall architecture of the proposed video inpainting localization network TruVIL. Uniformer blocks are employed to capture the inpainting traces both in RGB and noise streams, where the block numbers $\textit{L}_i=\{5, 8, 20, 7\}$. Multilayer perceptron (MLP) layer followed by an attentive noise decoding module are used as a decoder for generating the final binary localization map. (Best viewed in color.)}
\label{fig_5}
\end{figure*}

\subsection{Inpainting Forensics}
Many recent works have made attempts to learn effective forensic features for image/video manipulation localization, as shown in Table \ref{tab0}. We describe in brief how these attempts are implemented and explain the novelty of our scheme accordingly. 

In order to suppress content disturbance, Li \textit{et al.} \cite{li2019localization} propose to implement the first convolutional layer with trainable high-pass filters. Yang \textit{et al.} \cite{yang2020constrained} use BayarConv\cite{bayar2018constrained} as the initial convolutional layer of a forensic network. Although such constrained convolutional layers are helpful for extracting noise features, they fail to cover the forensic information from RGB modality. Therefore, an increasing number of forensic methods\cite{wu2021iid, zhou2018learning ,wu2019mantra, dong2022mvss, guillaro2023trufor, zhou2021deep, yu2021frequency, wei2022deep} rely on both RGB and noise modalities. Zhou \textit{et al.} \cite{zhou2018learning} develop a two-stream Faster R-CNN, which addresses the input RGB image and its noise counterpart generated by the spatial rich model (SRM)\cite{fridrich2012rich}. Yao \textit{et al.} \cite{yao2024deep} extracts time-series residual by SRM filters, which are then fed into the dual-stream network for feature learning. The trainable high-pass filters\cite{wei2022deep}, error level analysis (ELA)\cite{zhou2021deep} and frequency domain information\cite{yu2021frequency} are also exploited to capture features of the modality beyond RGB. We also use SRM filters to disclose the regions with inconsistent noise artifacts. Different from the prior approaches that solely extract noises from the input image, we take it a step further by applying SRM filters to the low-level features at multiple scales, resulting in richer and more informative noise features. 

The features of different modalities are typically fused via various strategies. The feature concatenation at an early stage is adopted by the forensic methods including IID-Net\cite{wu2021iid}, Mantra-Net\cite{wu2019mantra} and VIDNet\cite{zhou2021deep}, while that is enforced at a middle stage in FAST\cite{yu2021frequency}, STTNet \cite{wei2022deep}, VIFST \cite{pei2023vifst}, UVL \cite{pei2023uvl} and CDSNet \cite{yao2024deep}. In addition, TruFor\cite{guillaro2023trufor} fuses features at a middle stage by means of a cross-modal calibration module. MVSS-Net\cite{dong2022mvss} and RGB-N\cite{zhou2018learning} achieve the fusion at the late stage. Different from the non-trainable bilinear pooling used in RGB-N, the dual attention\cite{fu2019dual} fusion module in MVSS-Net is trainable and thus more selective. Unlike any of them, our TruVIL performs the feature fusion at multiple stages through three proposed modules to take full advantage of noise features, which will be discussed in Section III detailedly.

\section{Proposed Approach}
In this section, we propose the trusted video inpainting localization network TruVIL shown in Fig. 2. To fully exploit noise features, three functional modules are devised specifically, \textit{i.e.},\ a multi-scale noise extraction module, a cross-modality attentive fusion module, and an attentive noise decoding module. 

\subsection{Overall Framework}
Note that Uniformer\cite{li2023Uniformer} can model the long-range dependency across multiple frames and explore the local spatial features via self-attention mechanism. Therefore, Uniformer is modified as the backbone encoder network, which is composed of the RGB and noise branches. The lightweight MLP layer\cite{xie2021segformer} followed by the attentive noise decoding module is deployed as a decoder for generating the inpainting localization maps.

\begin{figure*}[!t]
\centering
\includegraphics[width=\textwidth]{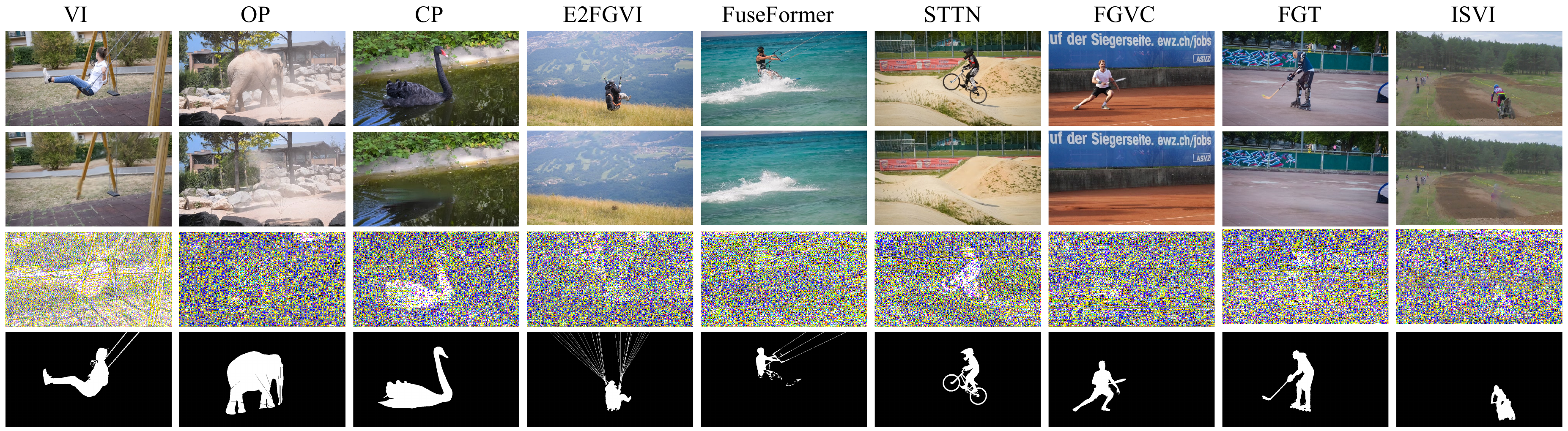}
\caption{Illustration of inpainting artifacts. From top to bottom: original frames, inpainted RGB frames and their corresponding noise images, and ground-truth inpainting masks. For different deep video inpainting algorithms, \textit{i.e.}, VI\cite{kim2020vipami}, OP\cite{oh2019onion}, CP\cite{lee2019copy}, E2FGVI\cite{li2022towards}, FuseFormer\cite{liu2021fuseformer}, STTN\cite{zeng2020learning}, FGVC\cite{gao2020flow}, FGT\cite{zhang2022flow} and ISVI\cite{zhang2022inertia}, the artifacts incurred by inpainting are hardly observed in the RGB space but clearly visible in the noise domain.}
\label{fig_1}
\end{figure*}

\begin{figure}[!t]
\centering
\includegraphics[width=\linewidth]{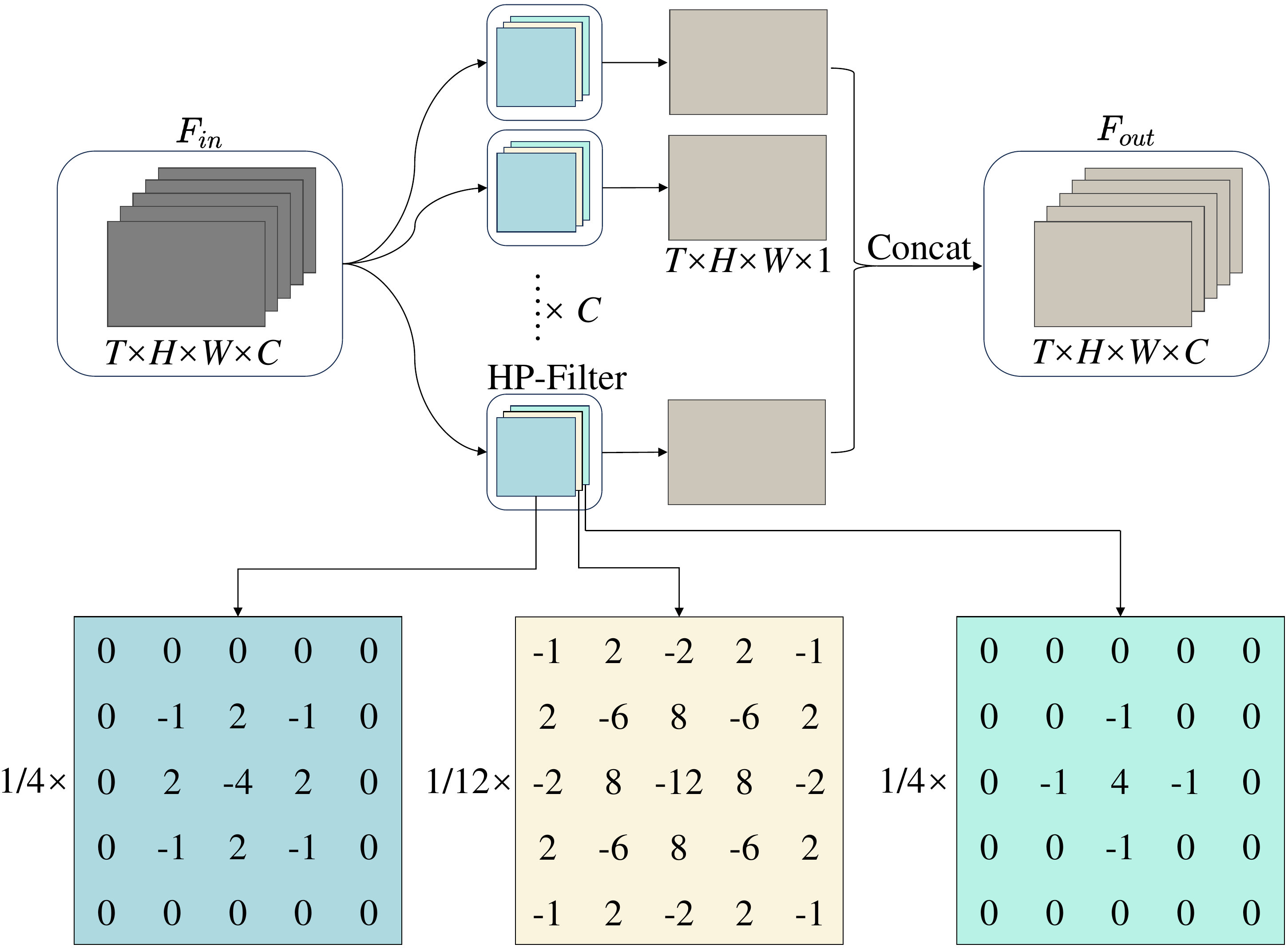}
\caption{Details of the 3D high pass filter (HP3D) layer. The HP-Filter consists of 3 convolution kernels with fixed parameters. $F_{in}$ and $F_{out}$ denote the input and output feature maps with the dimension $T \times H \times W \times C$, respectively.}

\label{fig_4}
\end{figure}

As shown in Fig. 2, given an inpainted video sequence $\textbf{\textit{F}}^0=\{f_{1}, f_{2}, \dotsb, f_{T}\}$ consisting of $\mathit{T}$ consecutive frames, where $f_i \in \mathbb{R}^{H \times W \times C}$. The TruVIL: $\mathbb{R}^{T \times H \times W \times C} \rightarrow \mathbb{R}^{H \times W \times 1}$ takes $\textbf{\textit{F}}^0$ as input, and outputs the predicted binary localization map. Specifically, the Uniformer-based encoder network transforms the $\textbf{\textit{F}}^0$ and its noise signal $\textbf{\textit{F}}_n^0$ to a series of multi-scale features $\{\textbf{\textit{F}}_i\}=\{\textbf{\textit{F}}_n^1, \textbf{\textit{F}}_n^2, \textbf{\textit{F}}_n^3, \tilde{\textbf{\textit{F}}}_n^4, \textbf{\textit{F}}^4\}$. Each Uniformer block consisting of Dynamic Position Embedding (DPE), Multi-Head Relation Aggregator (MHRA), and Feed-Forward Network (FFN) is illustrated in the down left of Fig. 2. For an input tensor $\textbf{\textit{X}}_{in}$, the DPE is first introduced to dynamically integrate 3D position information into all the tokens. Then the MHRA aggregates each token with its contextual ones. Finally, FFN with two linear layers is added for point-wise enhancement of each token. Such a Uniformer block can be formulated as 

{\setlength\abovedisplayskip{0.2cm}
\setlength\belowdisplayskip{0.2cm}
\begin{align}
\textbf{\textit{Y}}&=\operatorname{DPE}\left(\textbf{\textit{X}}_{in}\right)+\textbf{\textit{X}}_{in} \nonumber \\ 
\textbf{\textit{Z}}&=\operatorname{MHRA}(\operatorname{Norm}(\textbf{\textit{Y}}))+\textbf{\textit{Y}} \tag{1} \\ 
\textbf{\textit{X}}_{out}&=\operatorname{FFN}(\operatorname{Norm}(\textbf{\textit{Z}}))+\textbf{\textit{Z}} \nonumber
\end{align}}

\noindent 
where $\operatorname{Norm}(\cdot)$ is the normalization operator. 

The multi-scale features are then decoded by the MLP layer illustrated in the right down of Fig. 2. $\{\textbf{\textit{F}}_i\}=\{\textbf{\textit{F}}_n^1, \textbf{\textit{F}}_n^2, \textbf{\textit{F}}_n^3, \tilde{\textbf{\textit{F}}}_n^4, \textbf{\textit{F}}^4\}$ are first transformed by a linear layer for unifying the channel dimension, then upsampled to 1/4 size of $\textbf{\textit{F}}^0$ and channel-wise concatenated together. The resulting features are further fused by another linear layer for outputting the feature $\textbf{\textit{F}}$. Such MLP layer is formulated as

{\setlength\abovedisplayskip{0.2cm}
\setlength\belowdisplayskip{0.2cm}
\begin{align}
&\hat{\textbf{\textit{F}}}_\mathit{i} =\operatorname{Linear}\left(C_i, C\right)\left(\textbf{\textit{F}}_\mathit{i} \right), \forall \textbf{\textit{F}}_\mathit{i}  \nonumber \\
& \hat{\textbf{\textit{F}}}_\mathit{i} =\operatorname{Upsample}\left(\frac{H}{4} \times \frac{W}{4}\right)\left(\hat{\textbf{\textit{F}}}_\mathit{i} \right), \forall \hat{\textbf{\textit{F}}}_\mathit{i}  \tag{2} \\
& \textbf{\textit{F}}=\operatorname{Linear}(5C, C)\left(\operatorname{Concat}\left(\hat{\textbf{\textit{F}}}_\mathit{i} \right)\right), \forall \hat{\textbf{\textit{F}}}_\mathit{i}  \nonumber
\end{align}}

\noindent 
where $C_{\mathit{i}}$ is the channel number of $\textbf{\textit{F}}_\mathit{i}$, and $\text{Linear}\left(C_{\mathit{in}}, C_{\mathit{out}}\right)(\cdot)$ denotes a linear layer with $C_{\mathit{in}}$ and $C_{\mathit{out}}$ as input and output vector dimensions, respectively. Finally, the feature $\textbf{\textit{F}}$ is converted to a binary localization map by the attentive noise decoding module (See Part E of Section III). 

\subsection{Multi-scale Noise Extraction in Early Stage}
Noise is an intrinsic specificity of an image and can be found in various forms in all digital imagery domains. Since tampering operations destroy the consistency of noise distribution in original images, there often leave distinctive traces in noise space\cite{mahdian2009using}. In many image forensic methods\cite{li2019localization, wu2021iid, wu2019mantra, dong2022mvss}, a common practice for capturing tampering traces is to extract noise by high-pass filtering. Inspired by such works, we propose HP3D layers to uncover the inpainting traces. Fig. 4 shows details of a HP3D layer, which can be put in arbitrary positions or branches of our network. It includes adaptive number of HP-Filters, and each has 3 convolution kernels from the SRM. Given an input feature map $\textbf{\textit{F}}_{in} \in \mathbb{R}^{T \times H \times W \times C} $, its corresponding noise feature $\textbf{\textit{F}}_{out}$ is yielded by the HP3D layer as
\begin{align}
& \textbf{\textit{F}}_{out} = \operatorname{HP3D}(\textbf{\textit{F}}_{in}).  \tag{3}
\end{align}

As shown in Fig. 3, the artifacts left by inpainting algorithms can be observed clearly in the noise image yielded by the HP3D layer. In this way, the forgery areas can be detected in the noise domain.

\begin{figure*}[!t]
\centering
\includegraphics[width=\textwidth]{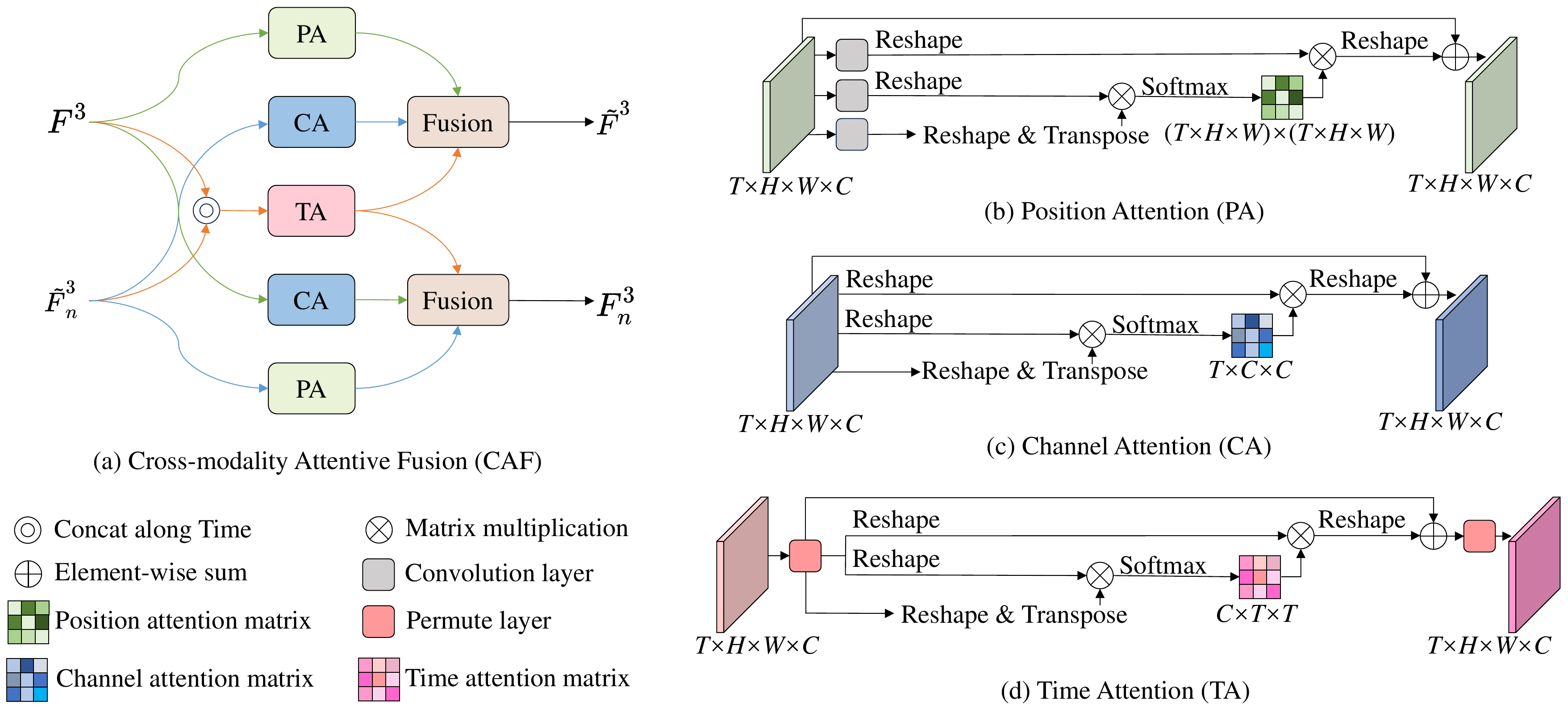}
\caption{Proposed cross modality attention module (Left) and its three constituent sub-modules (Right).}
\label{fig_6}
\end{figure*}

To enrich the noise features, we propose a multi-scale noise extraction (MNE) module which applies the HP3D layers to multiple low-level feature maps. As shown in Fig. 2, the input video sequence $\textbf{\textit{F}}^0 \in \mathbb{R}^{T \times H \times W \times C}$ is first converted to ist noise $\textbf{\textit{F}}_n^0 \in \mathbb{R}^{T \times H \times W \times C}$  by a HP3D layer. Then the two-stream Uniformer encoder takes $\textbf{\textit{F}}^0$ and $\textbf{\textit{F}}_n^0$ as inputs to generate two types of raw features, \textit{i.e.},\ the visual spatial feature maps $\textbf{\textit{F}}^i$ ($\mathit{i} =1, 2$) and the multi-scale noise feature maps $\textbf{\textit{F}}_\mathit{n} ^\mathit{i}$ ($\mathit{i} =1, 2$), respectively. That is, 
\begin{align}
& \textbf{\textit{F}}^\mathit{i}=\operatorname{Uniformer \ Block_\mathit{i} }\left(\textbf{\textit{F}}^{{i}-1}\right), \mathit{i} =1, 2 \nonumber \\
& \tilde{\textbf{\textit{F}}}_\mathit{n}^\mathit{i}=\operatorname{Uniformer \ Block_\mathit{i} }\left(\textbf{\textit{F}}_\mathit{n}^{{i}-1}\right), \mathit{i} =1, 2  \tag{4} \\
& \textbf{\textit{F}}_\mathit{n} ^\mathit{i}=\tilde{\textbf{\textit{F}}}_\mathit{n}^\mathit{i} + \operatorname{HP3D}(\textbf{\textit{F}}^\mathit{i}), \mathit{i} =1, 2. \nonumber
\end{align}

\subsection{Cross-modality Attentive Fusion in Middle Stage}
Attention mechanism\cite{fu2019dual, niu2021review} has been used broadly in natural language processing and computer vision. Inspired by these works, we devise an attention module to model the interaction between RGB and noise features. Our cross-modality attentive fusion (CAF) module is derived from the dual attention mechanism\cite{fu2019dual}, which consists of two types of attention modules, \textit{i.e.},\ Position Attention (PA) and Channel Attention (CA). PA selectively aggregates the feature at each position by a weighted sum of the features at all positions. And CA selectively emphasizes the interdependent channel maps by integrating the associated features among all channel maps. We adapt the PA and CA to videos by extending the dual attention from 2D image space to space-time 3D volume. 

Considering that consecutive multiple frames are used as inputs to obtain the localization map of the middle frame, we propose the novel Time Attention (TA) to aggregate inter-frame forensic information. As illustrated in Fig. 5(d), given an input feature map $\textbf{\textit{X}} \in \mathbb{R}^{T \times H \times W \times C}$, it is converted to $\mathbb{R}^{C \times T \times H \times W}$ by a permute layer, and then reshaped to $\mathbb{R}^{C \times T \times N}$, where $ N = H \times W $. After that matrix multiplication is performed between $\textbf{\textit{X}}$ and the transpose of $\textbf{\textit{X}}$. Finally, a softmax layer is enforced to generate the time attention map $\textbf{\textit{M}} \in \mathbb{R}^{C \times T \times T }$ as
\begin{align}
& m_{ji}=\frac{exp(\textbf{\textit{X}}_i\cdot \textbf{\textit{X}}_j)}{\sum_{i=1}^Texp(\textbf{\textit{X}}_i\cdot \textbf{\textit{X}}_j)} \tag{5}
\end{align}

\noindent 
where $m_{ji}$ measures the impact of the $i^{th}$ time dimension on the $j^{th}$ one. In addition, we perform a matrix multiplication between $\textbf{\textit{M}}$ and $\textbf{\textit{X}}$, and reshape their result to $ \mathbb{R}^{C \times T \times H \times W}$. Then we multiply the result by a scale parameter $\beta$ and perform an element-wise sum operation with $\textbf{\textit{X}}$ to obtain the output $\hat{\textbf{\textit{X}}} \in \mathbb{R}^{C \times T \times H \times W} $ as
\begin{align}
\label{eq6}
& \hat{\textbf{\textit{X}}}_j=\beta\sum_{i=1}^T(m_{ji}\textbf{\textit{X}}_i)+\textbf{\textit{X}}_j \tag{6}
\end{align}

\noindent 
where $\beta$ gradually learns a weight from 0. Finally, $\hat{\textbf{\textit{X}}}$ is converted to $ \mathbb{R}^{T \times H \times W \times C} $ by a permute layer.

Eq. \ref{eq6} shows that the final feature of each time dimension is a weighted sum of the features of all time dimensions and original features, which model long-range interdependencies among time dimensions. The PA and CA share similar processing process with TA, as shown in Fig. 5(b) and Fig. 5(c). Below the cross-modality attentive fusion of two branches features in the CAF module is described in detail.  

In the spatial dimension, PA is applied to $\textbf{\textit{F}}^3$ and $\tilde{\textbf{\textit{F}}}_n^3$ for getting the high-level position features $\textbf{\textit{F}}_p^1$  and $\textbf{\textit{F}}_p^2$. Meanwhile, CA is applied to $\textbf{\textit{F}}^3$ and $\tilde{\textbf{\textit{F}}}_n^3$ to get the high-level channel features $\textbf{\textit{F}}_c^1$ and $\textbf{\textit{F}}_c^2$. And in the temporal dimension, different from PA and CA, $\textbf{\textit{F}}^3$ and $\tilde{\textbf{\textit{F}}}_n^3$ are concatenated along time dimension, then fed into a TA module to get the high-level time feature $\textbf{\textit{F}}_t$. Finally, these high-level spatiotemporal features are cross fused to obtain $\tilde{\textbf{\textit{F}}}^\mathrm{3}$ and $\textbf{\textit{F}}_n^\mathrm{3}$. It can be formulated as
\begin{align}     
\textbf{\textit{F}}_t &= \operatorname{Conv}(\operatorname{TA}(\operatorname{CAT}(\textbf{\textit{F}}^3, \textbf{\textit{F}}_n^\mathrm{3}))) \nonumber \\
\tilde{\textbf{\textit{F}}}^\mathrm{3} &= \operatorname{Conv}(\operatorname{CAC}(\textbf{\textit{F}}_p^1,\textbf{\textit{F}}_c^2,\textbf{\textit{F}}_t)) \tag{7} \\
\textbf{\textit{F}}_n^\mathrm{3} &= \operatorname{Conv}(\operatorname{CAC}(\textbf{\textit{F}}_p^2,\textbf{\textit{F}}_c^1,\textbf{\textit{F}}_t)) \nonumber
\end{align}

\noindent 
where $\text{CAT}(\cdot)$ and $\text{CAC}(\cdot)$ refer to concatenating along time dimension and concatenating along channel dimension. As a result, any two positions with similar features can contribute mutual improvement regardless of their distance in spatiotemporal dimensions thanks to the use of CAF module.

\subsection{Attentive Noise Decoding in Late Stage}
Low-level features can improve accurate prediction on the boundaries and details, but it may lead to misclassification on other regions \cite{fu2020scene}. Therefore, we adopt attention mechanism to guide the selective application of noise information, thus obtaining better feature fusion. We propose a simple yet effective attentive noise decoding (AND) module that selectively enhances spatial details to improve the localization performance, as shown in Fig. 6.

\begin{figure}[!t]
\centering
\includegraphics[width=\linewidth]{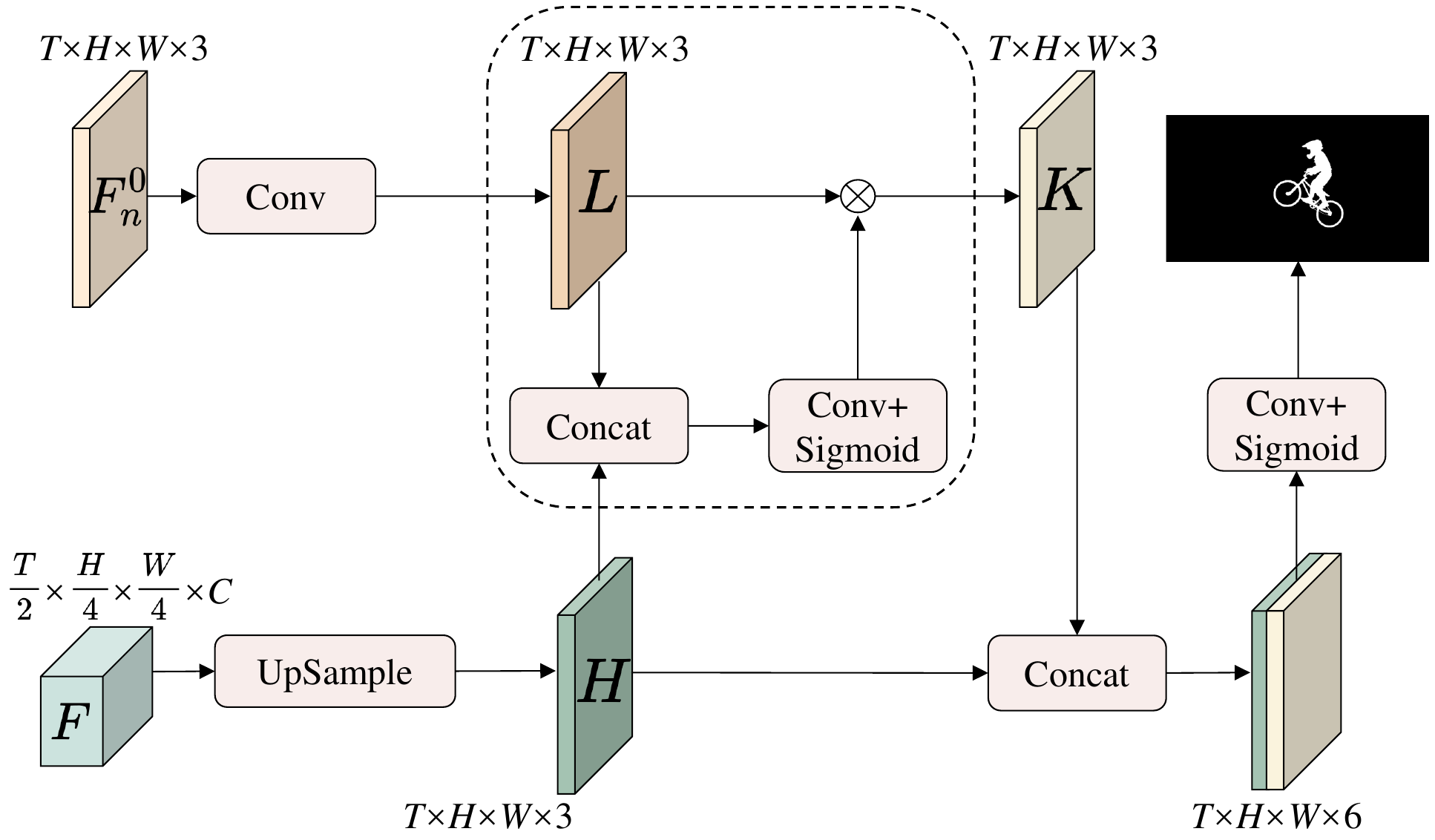}
\caption{Proposed attentive noise decoding module.}
\label{fig_7}
\end{figure}

First, a convolution layer is applied to $\textbf{\textit{F}}_n^0$ for yielding the low-level feature $\textbf{\textit{L}}$. Meanwhile, trilinear upsampling is applied to $\textbf{\textit{F}}$ for outputting the high-level feature $\textbf{\textit{H}}$. In order to fuse $\textbf{\textit{L}}$ and $\textbf{\textit{H}}$ effectively, a cross-level gate is deployed to refine them. The two features are concatenated and fed into a convolution layer followed by a sigmoid layer to obtain a spatial attention map. Then $\textbf{\textit{L}}$ is reweighted according to the attention map, making the region of details and boundaries more responsive. The refined low-level feature $\textbf{\textit{K}}$ is concatenated with $\textbf{\textit{H}}$ together. Finally, convolution layers followed by a sigmoid layer are configured for yielding the final binary localization map.

\subsection{Loss Function}
In practice, the inpainting algorithms are typically used to remove some salient video objects. Therefore, the inpainted regions are usually much smaller than the untouched ones. The standard cross entropy loss fails to handle such class imbalance, since it tends to focus on the majority of negative samples and leads to a low true positive rate. To attenuate such deficiency, we adopt focal loss\cite{lin2017focal}, which assigns an extra factor to the naive cross entropy term. As such, the gradient of different imbalanced samples can be controlled by the loss. The focal loss is defined as
\begin{align}
    \begin{split}
        \mathcal{L}_{\mathrm{Focal}}(y,\hat{y})=& \begin{aligned}-\sum\alpha\left(1-\hat{y}\right)^{\gamma}*y\log\left(\hat{y}\right)\end{aligned}  \\
        &\begin{aligned}-\sum(1-\alpha)\hat{y}^{\gamma}*(1-y)\log{(1-\hat{y})}\end{aligned}
    \end{split} \tag{8}
\end{align}

\noindent 
where $y \in \{0, 1\}$ denotes the pixel-level ground truth label and $\hat{y} \in [0, 1]$ denotes the corresponding prediction result. $\alpha$ and $\gamma$ are hyperparameters, which are empirically set as 0.5 and 2 respectively.

In addition, the metric of mean Intersection of Union (IoU) is used to foster more intersection between the prediction and ground-truth masks. The IoU loss function is defined as
\begin{align}
\mathcal{L}_{\mathrm{IoU}}(y,\hat{y})=1-\frac{\sum y*\hat{y}}{\sum(y+\hat{y}-y*\hat{y})+\epsilon} \tag{9}
\end{align}

\noindent 
where the hyperparameter $\epsilon$ is a small number for evading zero division.

Overall, the hybrid loss function for supervised training is defined as
\begin{align}
\mathcal{L}(y,\hat{y})=\lambda_1\mathcal{L}_{\mathrm{Focal}}(y,\hat{y})+\lambda_2\mathcal{L}_{\mathrm{IoU}}(y,\hat{y}) \tag{10}
\end{align}

\noindent 
where $\lambda_1$ and $\lambda_2$ are both set as 0.5. The loss functions $\mathcal{L}_{\mathrm{Focal}}$ and $\mathcal{L}_{\mathrm{IoU}}$ both play significant roles in the optimization. The focal loss assists to alleviate class imbalance and pay attention to hard samples. The IoU loss directly measures the evaluation metric and guides the network to predict inpainted regions more accurately.

\begin{table*}[!t]
\caption{Datasets used in our experiments. We indicate the inpainting datasets and their source datasets.}
\tabcolsep=3 pt
\label{tab-dataset}
\begin{adjustbox}{width=\textwidth}
\begin{tabular}{rccc|ccc}
\toprule[1pt]
\textbf{Source Dataset}     &\textbf{Videos}     & \textbf{Frames}   & \textbf{Applied Inpainting Algorithms}    & \textbf{Inpainting Dataset}  & \textbf{Videos}  & \textbf{Usage}        \\    
\midrule
VOS2k5                   & 800          & 102843                & VI, OP, CP        & \multirow{3}{*}{DVI2016} & \multirow{2}{*}{(800$+$30)$\times$3} & \multirow{2}{*}{train} \\
\multirow{2}{*}{DAVIS2016\cite{perazzi2016benchmark}} & 30    & 2079     & VI, OP, CP         &           &                 &                        \\
                                                      & 20    & 1376     & VI, OP, CP         &           & 20$\times$3              & test          \\ \midrule
\multirow{2}{*}{DAVIS2017\cite{Pont-Tuset_arXiv_2017}} & \multirow{2}{*}{90} & \multirow{2}{*}{6208} & \multirow{2}{*}{E2FGVI, FuseFormer, STTN, FGT, FGVC, ISVI}  & \multirow{2}{*}{DVI2017} & \multirow{2}{*}{90$\times$6}       & \multirow{2}{*}{test}  \\
                           &                     &                       &                                                        &                          &                             &                        \\
MOSE\cite{MOSE}                       & 100                 & 8183                  & E2FGVI, FuseFormer, STTN                               & MOSE100                  & 100$\times$3                      & test                  \\
\bottomrule[1pt]
\end{tabular}
\end{adjustbox}
\end{table*}

\section{Experimental Results}
In this section, extensive experiments are performed to evaluate our proposed video inpainting localization scheme on various benchmark scenarios. We first introduce detailed experimental settings, the test results and discussions followed by. The results compared with state-of-the-art methods are also presented. Finally, extensive ablation experiments further validate the effectiveness of our proposed modules.

\subsection{Setup}

\subsubsection{Datasets}
The following datasets are used in our experiments, details are shown in Table \ref{tab-dataset}.
\begin{itemize}
\item{VOS2k5: We collect 2500 videos from YouTube-VOS\cite{Xu_2018_ECCV} and GOT10k\cite{8922619} to construct a frame-level video object segmentation dataset with pixel-level annotation for every frame. Specifically, we use the video object segmentation algorithm Xmem\cite{cheng2022xmem} to obtain pixel-level annotations for all frames.}

\item{DAVIS2016: DAVIS2016\cite{perazzi2016benchmark} is one of the most famous benchmarks for deep video inpainting, which consists of 30 videos for training and 20 videos for testing. To prepare enough training samples, 800 videos chosen from VOS2k5 are added to the training set of DAVIS2016. DVI2016 dataset is generated by applying three state-of-the-art video inpainting algorithms — VI\cite{kim2020vipami}, OP\cite{oh2019onion} and CP\cite{lee2019copy} to DAVIS2016, regarding the ground truth mask as reference. Two of the three inpainted DVI2016 datasets are used for both training and in-domain testing. After that, we conduct additional cross-domain testing using the remaining inpainted dataset.}

\item{DAVIS2017 and MOSE: To assess the generalization of TruVIL on more datasets and inpainting algorithms, DVI2017 dataset is yielded by applying another six video inpainting algorithms — E2FGVI\cite{li2022towards}, FuseFormer\cite{liu2021fuseformer}, STTN\cite{zeng2020learning}, FGVC\cite{gao2020flow}, FGT\cite{zhang2022flow} and ISVI\cite{zhang2022inertia} to DAVIS2017\cite{Pont-Tuset_arXiv_2017}. In addition, 100 videos are collected from MOSE\cite{MOSE}, which is a video object segmentation dataset with complex scenes and crowded objects. The corresponding MOSE100 dataset is created by E2FGVI\cite{li2022towards}, FuseFormer\cite{liu2021fuseformer} and STTN\cite{zeng2020learning} .} 
\end{itemize}

\subsubsection{Evaluation Metrics} F1 and IoU are used as the metrics of pixel-level localization accuracy. For calculating F1 and IoU, threshold is necessary as the direct outputs of the network are probability values. In line with existing works, the threshold is set to 0.5 by default.

\subsubsection{Implementation Details} 
The proposed method is implemented using the PyTorch deep learning framework and adopting the AdamW as the optimizer. We train the network on a single A800 GPU with an initial learning rate $5 \times 10^{-4}$, which decays to $5 \times 10^{-6}$ in 10 epochs with cosine annealing strategy. Each of 5 consecutive frames are set as an input unit and the batch size is 8. All the frames used in training are resized to $432 \times 240$ pixels. The training process includes two phases. First, the network is trained for 25 epochs without any data enhancement. Then one quarter of the training set is compressed by H.264 with the Constant Rate Factor (CRF) 23, and the network is trained for 5 epochs in this stage. Such two-stage training greatly improves the robustness of TruVIL.

\subsubsection{Compared Methods}
For a fair and reproducible comparison, we have to be selective, choosing the state-of-the-art that meets one of the following three criteria: 1) pre-trained models released by paper authors, 2) source code publicly available, or 3) experiment results available in papers. Accordingly, we choose several published methods as follows:

\begin{itemize}
\item{Models available: Mantra-Net\cite{wu2019mantra}, MVSS-Net\cite{dong2022mvss}, IF-OSN\cite{wu2022robust}, TruFor\cite{guillaro2023trufor}, FOCAL\cite{wu2023rethinking} and IID-Net\cite{wu2021iid}. We use these models directly.}

\item{Code available: Videofact\cite{nguyen2022videofact}, which is trained using author-provided code. We cite its results where appropriate and use our re-trained model only when necessary.}

\item{Results available: NOI\cite{mahdian2009using}, CFA\cite{ferrara2012image}, COSNet\cite{lu2019see}, HP-FCN\cite{li2019localization}, GSR-Net\cite{zhou2020generate}, VIDNet\cite{zhou2021deep}, STTNet\cite{wei2022deep}, FAST\cite{yu2021frequency} and UVL \cite{pei2023uvl}. All the methods are evaluated in the same test dataset, and we cite their results directly.}
\end{itemize}

\subsection{Compared with State-of-the-Art Methods}
We first compare the performance of TruVIL with several related methods on the DVI2016 test dataset. The related methods include the state-of-the-art video inpainting detection methods consisting of VIDNet \cite{zhou2021deep}, STTNet \cite{wei2022deep}, FAST \cite{yu2021frequency} and UVL \cite{pei2023uvl}, the video segmentation method COSNet \cite{lu2019see}, and the image manipulation detection methods consisting of NOI \cite{mahdian2009using}, CFA \cite{ferrara2012image}, HP-FCN \cite{li2019localization} and GSR-Net \cite{zhou2020generate}. To explore the effect of different inpainting algorithms, all the models are trained on two video inpainting methods and tested on the other one.

Table \ref{tab1} shows the quantitative comparison results of F1 and IoU (higher are better). For all the three training settings, TruVIL outperforms other approaches on most trained video inpainting approaches. It presents the advantages of our approach to acquire inpainting artifacts distributed in the videos. Furthermore, TruVIL also exceeds other approaches on all the unseen video inpainting approaches. It presents the powerful generalization ability of our approach. For example, when trained on VI and CP methods, the IoU and F1 on OP associated with VIDNet and FAST only reach about $0.2$ and $0.3$. In contrast, our proposed method still has high accuracy, its IoU and F1 reach 0.53 and 0.67, respectively. Despite the distinct differences in inpainting traces left by OP compared to VI and CP, TruVIL successfully captures the common artifacts of the three inpainting algorithms through the deep utilization of noise features.

\begin{table*}[!t]
\centering
\caption{Accuracy comparison of different inpainting localization methods on DVI2016 dataset. All methods are trained on the datasets inpainted by VI and OP, OP and CP, VI and CP algorithms, respectively (denoted as ‘*’). ‘-’ denotes that the result is not available. Bold numbers represent the best results.}
\label{tab1}
\begin{adjustbox}{width=\textwidth}
\begin{tabular}{rccccccccc}
\toprule[1pt]
              & \textbf{VI*}       & \textbf{OP*}       & \textbf{CP}                             & \textbf{VI}        & \textbf{OP*}       & \textbf{CP*}                            & \textbf{VI*}       & \textbf{OP}        & \textbf{CP*}       \\
Methods       & IoU/F1             & IoU/F1             & \multicolumn{1}{c|}{IoU/F1}             & IoU/F1             & IoU/F1             & \multicolumn{1}{c|}{IoU/F1}             & IoU/F1             & IoU/F1             & IoU/F1             \\ 
\midrule
NOI{ \cite{mahdian2009using}}     & 0.08/0.14          & 0.09/0.14          & \multicolumn{1}{c|}{0.07/0.13}          & 0.08/0.14          & 0.09/0.14          & \multicolumn{1}{c|}{0.07/0.13}          & 0.08/0.14          & 0.09/0.14          & 0.07/0.13          \\
CFA{ \cite{ferrara2012image}}     & 0.10/0.14          & 0.08/0.14          & \multicolumn{1}{c|}{0.08/0.12}          & 0.10/0.14          & 0.08/0.14          & \multicolumn{1}{c|}{0.08/0.12}          & 0.10/0.14          & 0.08/0.14          & 0.08/0.12          \\
COSNet{ \cite{lu2019see}}  & 0.40/0.48          & 0.31/0.38          & \multicolumn{1}{c|}{0.36/0.45}          & 0.28/0.37          & 0.27/0.35          & \multicolumn{1}{c|}{0.38/0.46}          & 0.46/0.55          & 0.14/0.26          & 0.44/0.53          \\
HP-FCN{ \cite{li2019localization}}     & 0.46/0.57          & 0.49/0.62          & \multicolumn{1}{c|}{0.46/0.58}          & 0.34/0.44          & 0.41/0.51          & \multicolumn{1}{c|}{0.68/0.77}          & 0.55/0.67          & 0.19/0.29          & 0.69/0.80          \\
GSR-Net{ \cite{zhou2020generate}} & 0.57/0.69          & 0.50/0.63          & \multicolumn{1}{c|}{0.51/0.63}          & 0.30/0.43          & 0.74/0.82          & \multicolumn{1}{c|}{0.80/0.85}          & 0.59/0.70          & 0.22/0.33          & 0.70/0.77          \\
VIDNet{ \cite{zhou2021deep}}  & 0.59/0.70          & 0.59/0.71          & \multicolumn{1}{c|}{0.57/0.69}          & 0.39/0.49          & 0.74/0.82          & \multicolumn{1}{c|}{0.81/0.87}          & 0.59/0.71          & 0.25/0.34          & 0.76/0.85          \\
STTNet{ \cite{wei2022deep}}  & 0.60/\textbf{0.73} & 0.69/0.80          & \multicolumn{1}{c|}{0.65/0.77}          & -                  & -                  & \multicolumn{1}{c|}{-}                  & -                  & -                  & -                  \\
FAST{ \cite{yu2021frequency}}    & 0.61/\textbf{0.73} & 0.65/0.78          & \multicolumn{1}{c|}{0.63/0.76}          & 0.32/0.49          & 0.78/0.87          & \multicolumn{1}{c|}{\textbf{0.82/0.90}} & 0.57/0.68          & 0.22/0.34          & 0.76/0.83          \\
UVL{ \cite{pei2023uvl}}    & \;\textbf{0.65}/\,\; - \;\; & \;0.66/\,\; - \;\;          & \multicolumn{1}{c|}{\;0.65/\,\; - \;\;}     & \textbf{\;0.64}/\,\; - \;\;  & \;0.67/\,\; - \;\;  & \multicolumn{1}{c|}{\;0.68/\,\; - \;\;} & \textbf{\;0.75}/\,\; - \;\;          & \textbf{\;0.75}/\,\; - \;\;          & \;0.74/\,\; - \;\;          \\
TruVIL (ours)          & 0.61/0.72 & \textbf{0.82/0.89} & \multicolumn{1}{c|}{\textbf{0.70/0.81}} & 0.42/\textbf{0.54} & \textbf{0.84/0.91} & \multicolumn{1}{c|}{\textbf{0.82}/0.89} & 0.63/\textbf{0.74} & 0.53/\textbf{0.67} & \textbf{0.81/0.88} \\ 
\bottomrule[1pt]
\end{tabular}
\end{adjustbox}
\end{table*}

\begin{table*}[!t]
\centering
\caption{Accuracy comparison of different inpainting localization methods on DVI2017 dataset. Our model is trained on DVI2016 dataset inpainted by VI and OP algorithms and directly tested on DVI2017 dataset. ‘-’ denotes that the result is not available. Bold numbers represent the best results.}

\label{tab2}
\begin{adjustbox}{width=\textwidth}
\begin{tabular}{rccccccc}
\toprule[1pt]
                                        & \textbf{E2FGVI}{\cite{li2022towards}} & \textbf{FuseFormer} {\cite{liu2021fuseformer}}    & \textbf{STTN} {\cite{zeng2020learning}}          & \textbf{FGT} {\cite{zhang2022flow}}           & \textbf{FGVC} {\cite{gao2020flow}}          & \textbf{ISVI} {\cite{zhang2022inertia}}     & \textbf{Average}    \\
Methods                                 & IoU/F1                & IoU/F1                     & IoU/F1               & IoU/F1                & IoU/F1                & IoU/F1        & IoU/F1       \\
\midrule
Mantra-Net\cite{wu2019mantra}        & 0.299/0.426               & 0.401/0.537                & 0.334/0.475          & 0.207/0.318           & 0.244/0.367          & 0.275/0.394     & 0.293/0.420  \\
MVSS-Net\cite{dong2022mvss}          & 0.030/0.047               & 0.048/0.069                & 0.096/0.136          & 0.045/0.070           & 0.054/0.083          & 0.082/0.117     & 0.059/0.087  \\
IF-OSN\cite{wu2022robust}            & 0.041/0.065               & 0.044/0.069                & 0.032/0.051          & 0.029/0.050           & 0.027/0.045          & 0.100/0.147     & 0.046/0.071  \\
TruFor\cite{guillaro2023trufor}      & 0.231/0.318               & 0.211/0.293                & 0.166/0.233          & 0.234/0.325           & 0.266/0.367          & 0.341/0.443     & 0.242/0.330  \\
FOCAL\cite{wu2023rethinking}         & 0.088/0.134               & 0.142/0.210                & 0.147/0.218          & 0.103/0.160           & 0.082/0.131          & 0.337/0.430     & 0.150/0.214  \\
IID-Net\cite{wu2021iid}              & 0.192/0.285               & 0.210/0.303                & 0.192/0.284          & 0.120/0.194           & 0.096/0.163          & 0.119/0.196     & 0.155/0.238  \\
Videofact\cite{nguyen2022videofact}  & \quad - \enspace /0.309   & \quad - \enspace /0.237    & 0.097/0.261          & 0.082/0.246           & 0.073/0.248          & 0.037/0.220     & 0.072/0.254  \\
TruVIL (ours)                        & \textbf{0.572/0.697}      & \textbf{0.644/0.762}       & \textbf{0.589/0.715} & \textbf{0.455/0.592}  & \textbf{0.396/0.533} & \textbf{0.344/0.470} & \textbf{0.500/0.628} \\
\bottomrule[1pt]
\end{tabular}
\end{adjustbox}
\end{table*}

\begin{table}[!t]
\centering
\caption{Accuracy comparison of different inpainting localization methods on MOSE100 dataset. Our model is trained on DVI2016 dataset inpainted by VI and OP algorithms and directly tested on MOSE100 dataset. Bold numbers represent the best results.}
\tabcolsep=3 pt
\label{tab3}
\begin{adjustbox}{width=\linewidth}
\begin{tabular}{rcccccc}
\toprule[1pt]
                                           & \textbf{E2FGVI} {\cite{li2022towards}}        & \textbf{FuseFormer} {\cite{liu2021fuseformer}}    & \textbf{STTN} {\cite{zeng2020learning}}   \\
Methods                                    & IoU /  F1               & IoU /  F1               & IoU /  F1               \\
\midrule
Mantra-Net {\cite{wu2019mantra}}                & 0.378/0.524          & 0.385/0.531          & 0.356/0.505          \\
MVSS-Net {\cite{dong2022mvss}}                  & 0.038/0.057          & 0.051/0.074          & 0.094/0.133          \\
IF-OSN {\cite{wu2022robust}}                   & 0.041/0.068          & 0.041/0.067          & 0.031/0.050          \\
TruFor {\cite{guillaro2023trufor}}          & 0.311/0.414          & 0.285/0.388          & 0.260/0.353          \\
FOCAL {\cite{wu2023rethinking}}             & 0.098/0.150          & 0.138/0.206          & 0.152/0.226          \\
IID-Net {\cite{wu2021iid}}                  & 0.084/0.140          & 0.087/0.143          & 0.082/0.137          \\
Videofact {\cite{nguyen2022videofact}}      & 0.085/0.187          & 0.076/0.177          & 0.088/0.191           \\
TruVIL (ours)                             & \textbf{0.521/0.674} & \textbf{0.557/0.699} & \textbf{0.462/0.612} \\
\bottomrule[1pt]
\end{tabular}
\end{adjustbox}
\end{table}

\subsection{Generalization Analysis}
In this experiment, we compare the generalization performance of TruVIL and 7 existing methods. These methods includes 5 universal image manipulation detection methods consisting of Mantra-Net\cite{wu2019mantra}, MVSS-Net\cite{dong2022mvss}, IF-OSN\cite{wu2022robust}, TruFor\cite{guillaro2023trufor} and FOCAL\cite{wu2023rethinking}, a deep image inpainting detection method IID-Net\cite{wu2021iid}, and a deep video forgery detection method Videofact\cite{nguyen2022videofact}. We conduct generalization experiments in cross-algorithm and cross-dataset scenarios, where TruVIL is trained on DVI2016 dataset inpainted by VI and OP algorithms.

\begin{table*}[!t]
\caption{Accuracy comparison of different inpainting localization methods on compressed DVI2017 dataset. Our model is trained on DVI2016 dataset inpainted by VI and OP algorithms and directly tested on compressed DVI2017 dataset with different CRFs. Bold numbers represent the best results.}
\tabcolsep=2.5 pt
\label{tab4}
\begin{adjustbox}{width=\textwidth}
\begin{tabular}{rccccccccc}

\toprule[1pt]
             & \multicolumn{3}{c}{\textbf{E2FGVI} {\cite{li2022towards}}}            & \multicolumn{3}{c}{\textbf{FuseFormer} {\cite{liu2021fuseformer}}}            & \multicolumn{3}{c}{\textbf{STTN} {\cite{zeng2020learning}}}            \\ 
             & CRF=18          & CRF=23      & \multicolumn{1}{c|}{CRF=28}     & CRF=18      & CRF=23      & \multicolumn{1}{c|}{CRF=28}       & CRF=18     & CRF=23      & CRF=28                \\
Methods      & IoU/F1       & IoU/F1      & \multicolumn{1}{c|}{IoU/F1}      & IoU/F1     & IoU/F1     & \multicolumn{1}{c|}{IoU/F1}      & IoU/F1       & IoU/F1         & IoU/F1               \\
\midrule
Mantra-Net {\cite{wu2019mantra}}               & 0.030/0.055 & 0.035/0.063 & \multicolumn{1}{c|}{0.037/0.068} & 0.048/0.086  & 0.051/0.090  & \multicolumn{1}{c|}{0.054/0.095} & 0.117/0.186  & 0.055/0.156  & 0.071/0.123  \\
MVSS-Net {\cite{dong2022mvss}}                 & 0.011/0.176 & 0.010/0.161 & \multicolumn{1}{c|}{0.014/0.238} & 0.019/0.029  & 0.014/0.022  & \multicolumn{1}{c|}{0.015/0.024} & 0.045/0.065  & 0.034/0.052  & 0.026/0.041  \\
IF-OSN {\cite{wu2022robust}}                  & 0.029/0.048 & 0.030/0.050 & \multicolumn{1}{c|}{0.027/0.044} & 0.025/0.043  & 0.024/0.042  & \multicolumn{1}{c|}{0.023/0.039} & 0.024/0.041  & 0.029/0.048  & 0.028/0.047  \\
TruFor {\cite{guillaro2023trufor}}         & 0.024/0.036 & 0.018/0.028 & \multicolumn{1}{c|}{0.019/0.031} & 0.034/0.052  & 0.025/0.040  & \multicolumn{1}{c|}{0.026/0.042} & 0.037/0.054  & 0.034/0.052  & 0.032/0.050  \\
FOCAL {\cite{wu2023rethinking}}            & 0.051/0.083 & 0.047/0.079 & \multicolumn{1}{c|}{0.041/0.071} & 0.060/0.098  & 0.056/0.092  & \multicolumn{1}{c|}{0.045/0.077} & 0.064/0.103  & 0.059/0.097  & 0.050/0.084  \\
IID-Net {\cite{wu2021iid}}                 & 0.112/0.188 & 0.115/0.192 & \multicolumn{1}{c|}{0.116/0.193} & 0.114/0.191  & 0.116/0.192  & \multicolumn{1}{c|}{0.117/0.193} & 0.107/0.181  & 0.113/0.189  & 0.115/0.191  \\
Videofact {\cite{nguyen2022videofact}}     & 0.093/0.247 & 0.087/0.243 & \multicolumn{1}{c|}{0.081/0.237} & 0.093/0.246 & 0.091/0.246  & \multicolumn{1}{c|}{0.090/0.242} & 0.118/0.271  & 0.114/0.268  & 0.107/0.259  \\
TruVIL (ours)     & \textbf{0.570/0.693} & \textbf{0.546/0.671} & \multicolumn{1}{c|}{\textbf{0.485/0.610}} & \textbf{0.635/0.751} & \textbf{0.599/0.720} & \multicolumn{1}{c|}{\textbf{0.525/0.653}} & \textbf{0.559/0.684} & \textbf{0.558/0.682} & \textbf{0.517/0.644} \\
\bottomrule[1pt]
\end{tabular}
\end{adjustbox}
\end{table*}

\begin{figure*}[!t]
\centering
\includegraphics[width=\textwidth]{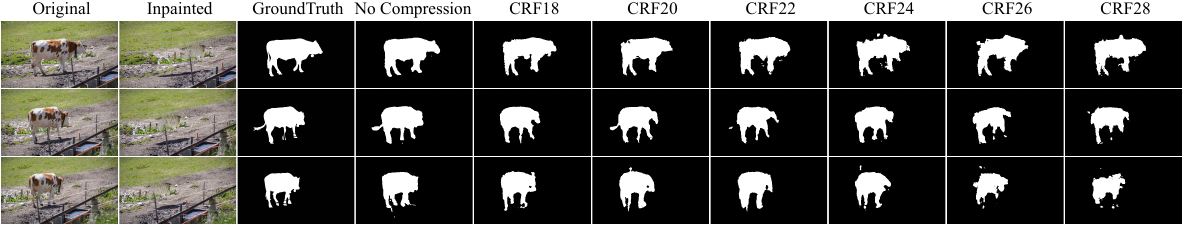}
\caption{Localization results on an example DVI2017 video compressed with different CRFs.}
\label{fig10}
\end{figure*}

\subsubsection{Cross-algorithm }
Generalization experiments are first performed on DVI2017 dataset, which is generated using another six deep video inpainting algorithms. As shown in Table \ref{tab2}, all the approaches suffer from the performance degradation in cross-algorithm scenario. However, TruVIL still achieves better generalization compared to existing methods. For instance, the F1 on E2FGVI and FuseFormer associated with Videofact only reach 0.247 and 0.242. On the contrary, our proposed method reliably achieves precise inpainting localization, as demonstrated by the F1 of 0.697 and 0.762, respectively. The average F1 and IoU of TruVIL on these six inpainting algorithms also far outperform other existing forensic methods. Although there are significant differences in various inpainting algorithms, they leave similar artifacts in the noise domain. TruVIL exhibits excellent generalization ability thanks to the utilization of noise features. 

\subsubsection{Cross-dataset}
To further evaluate the generalization ability in cross-dataset scenario, we test TruVIL on MOSE100 dataset. As shown in Table \ref{tab3}, our method outperforms the existing ones on all inpainted datasets. For the best baseline, \textit{i.e.},\ Mantra-Net, its F1 only reaches about 0.5. As for TruVIL, its F1 is consistently higher than 0.6, which again justifies the excellent generalization ability of our model.

\subsection{Robustness Evaluation}
We would also like to evaluate the robustness of TruVIL in deep video inpainting localization. It is highly crucial in real-world forensic scenarios, as numerous post-processing may be employed to weaken the forgery traces. Previous works\cite{zhou2021deep, yu2021frequency} studied the robustness of video forensic methods against JPEG and Gaussian noise perturbation. However, the post-processing that videos are more likely to encounter is video compression. Therefore, it is more reasonable to evaluate the robustness against video compression.

\begin{figure*}[p]
\centering
\includegraphics[width=\textwidth]{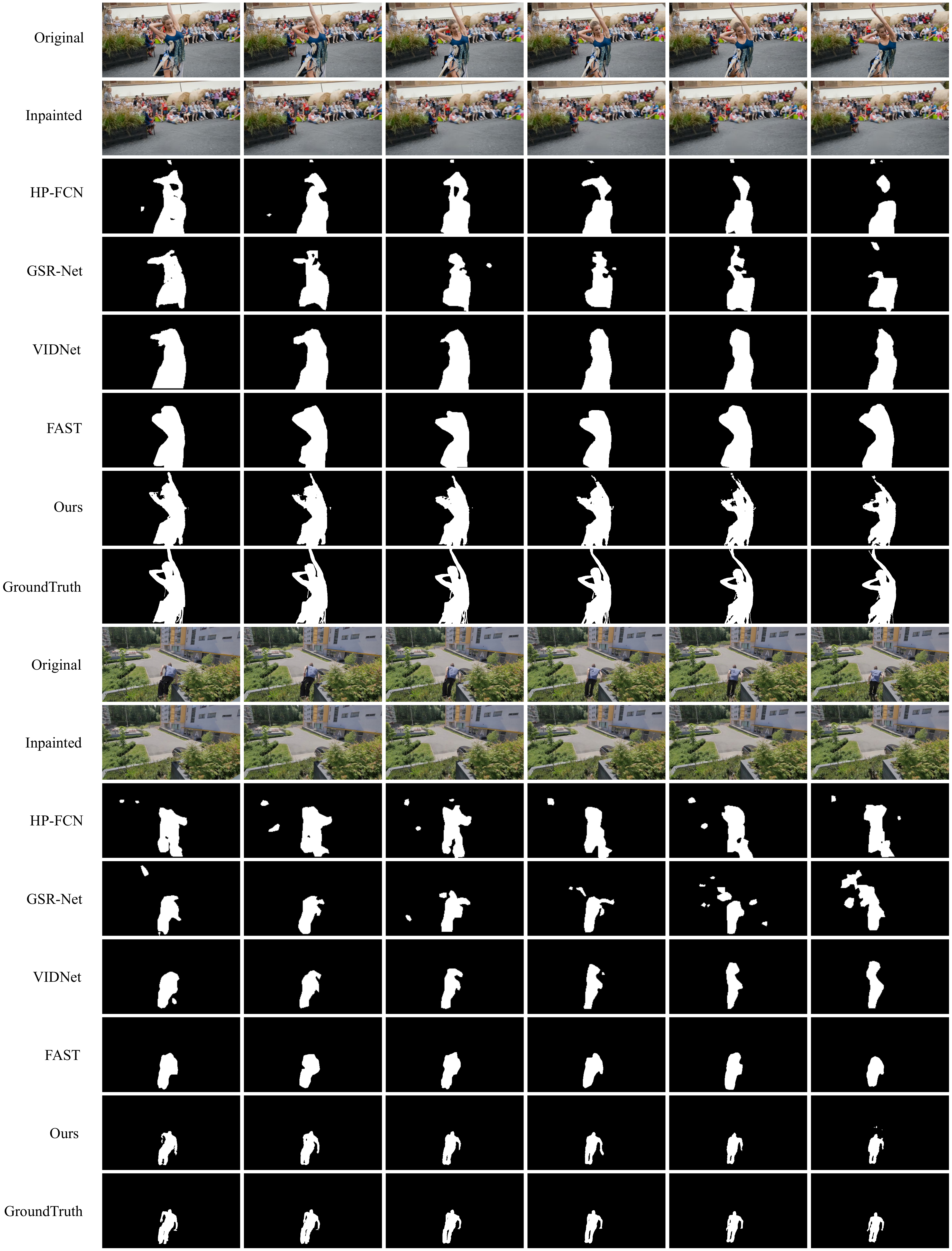}
\caption{Qualitative visualization on two DVI2016 videos. The first and second rows show the original and inpainted frames. The third to seventh row indicates the final predictions from different methods. The eighth row is the ground truth.}
\label{fig_11}
\end{figure*}

CRF is a parameter for controlling H.264 compression quality, and typically ranges from 0 to 51, where 0 represents lossless compression, 51 represents the lowest quality compression, and 23 is the default value. We test our model on compressed DVI2017 dataset with different CRFs, where TruVIL is trained on DVI2016 dataset inpainted by VI and OP algorithms. 

As illustrated in Table \ref{tab4}, compared to the uncompressed scenario (See Table \ref{tab2}), the localization performance of TruVIL shows a small drop under H.264 compression. For example, when CRF is from 18 to 28, our method only experiences a decrease in IoU on the STTN inpainted dataset from 0.559 to 0.517, and a decrease in F1 from 0.684 to 0.644. In particular, our F1 values are consistently higher than 0.6 in all compression cases. Fig. 7 shows the localization results on a DVI2017 video compressed with different CRFs. It indicates that TruVIL demonstrates excellent robustness against video compression of high and medium magnitudes. Nevertheless, as the intensity of compression escalates, the integrity of the inpainting evidence is compromised, resulting in detection errors. However, strong compression also lead to severely degraded videos, which deviates the purpose of performing inpainting. On the whole, our two-stage robust training strategy could be a viable solution for improving the robustness.

\begin{table*}[!t]
\centering
\caption{Evaluation of different components of our proposed approach. The models are trained on DVI2016 dataset inpainted by VI and OP algorithms and tested on DVI2016 dataset inpainted by CP.}
\tabcolsep=2.5 pt
\label{tab5}
\begin{adjustbox}{width=\textwidth}
\begin{tabular}{ll|ccccccccc}
\toprule[1pt]
\multirow{3}{*}{Modality} & RGB                                          & \checkmark &            &            &            &            &            &            &            &                \\
                          & Noise                                        &            & \checkmark &            &            &            &            &            &            &                \\
                          & RGB+Noise                                    &            &            & \checkmark & \checkmark & \checkmark & \checkmark & \checkmark & \checkmark & \checkmark     \\ 
\midrule
\multirow{4}{*}{Module}   & w/o MNE                                      &            &            &            & \checkmark &            &            &            &            &                \\
                          & w/o CAF                                      &            &            &            &            & \checkmark &            &            &            &                \\
                          & w/o AND                                      &            &            &            &            &            & \checkmark &            &            &                \\
                          & MNE+CAF+AND                                  &            &            &            &            &            &            & \checkmark & \checkmark & \checkmark     \\ 
\midrule
\multirow{3}{*}{Loss}     & $\mathcal{L}_{\mathrm{Focal}}$                         &            &            &            &            &            &            & \checkmark &            &                \\
                          & $\mathcal{L}_{\mathrm{IoU}}$                           &            &            &            &            &            &            &            & \checkmark &                \\
                          & $\mathcal{L}_{\mathrm{Focal}}$ + $\mathcal{L}_{\mathrm{IoU}}$    & \checkmark & \checkmark & \checkmark & \checkmark & \checkmark & \checkmark &            &            & \checkmark     \\ 
\midrule
Metric                    & IoU/F1           & 0.457/0.568  & 0.539/0.627  & 0.552/0.645  &  0.608/0.724 &  0.588/0.681 & 0.627/0.754  & 0.652/0.778  & 0.671/0.784  & 0.697/0.805 \\
\bottomrule[1pt]
\end{tabular}
\end{adjustbox}
\end{table*}

\subsection{Qualitative Results}
In addition to the quantitative comparisons, we also compare different methods qualitatively. Fig. 8 illustrates the visualization of our predictions compared with others under the same setting. It is observed that TruVIL predicts the masks closest to the ground truth. Specifically, HP-FCN tends to misclassify genuine regions due to the inherent constraints imposed by a singular input modality. Furthermore, the frame-by-frame inpainting detection is employed by GSR-Net, thereby compromising the coherence of the outputs. Though VIDNet achieves temporally consistent predictions by convolutional LSTM, its localization results exhibit occasional omissions of intricate details. Based on the extraction of frequency-aware features, FAST greatly improves the localization performance in spatial details, but it is prone to false alarms. Compared with these methods, our proposed TruVIL generates more precise predicted masks, which primarily thanks to the full utilization of attentive noise features and the carefully designed modules.

\subsection{Ablation Studies}
We conduct extensive ablation studies to analyze how each component of our model contributes to the final localization results. To this end, we prohibit the use of additional components in each network architecture, and then evaluate the performance of different re-trained models. Specifically, we evaluate proposed model and its variants on DVI2016 dataset, where models are trained on VI and OP methods, and tested on CP method. The results are shown in Table \ref{tab5}.

First, we only input RGB frames or noise features to a single stream network to conduct experiments. The results show that the noise input lead to better performance than RGB. This is mainly because noise features can unveil the inpainting traces, providing powerful evidence for inpainting localization. In addition, the localization performance is better when RGB frames and noise features are simultaneously used as inputs to form a two stream network.

Then, to demonstrate the benefit of designed modules, we evaluate each variant of our model, where one of the modules is removed. It can be seen that each of proposed modules can bring positive improvements. Specifically, our method without CAF module experiences a decrease in F1 from 0.805 to 0.681. It illustrates the importance of cross-modality attentive fusion, which can further optimize high-level features for boosting the eventual inpainting localization performance. Moreover, our method without MNE (only apply HP3D layer to the input) or AND module also experiences performance degradation. It shows that making full use of the noise features is beneficial to improve localization performance.

Finally, we drop the focal or IoU loss in the hybrid loss function. We can observe that the performance of our model deteriorates if either loss function is missing. It suggests that these two loss functions both play important roles in optimizing the TruVIL. In addition, the IoU loss $\mathcal{L}_{\mathrm{IoU}}$ is slightly more important because it is directly related to evaluation metrics.

\section{Conclusion}
In this paper, we present a trusted video inpainting localization network TruVIL based on the dual stream Uniformer encoder and MLP decoder. Our scheme leverages deep attentive noise features to reveal the inpainting traces. To fully exploit noise features, we carefully devised three novel modules including the multi-scale noise extraction, the cross-modality attentive fusion, and the attentive noise decoding. Such modules benefit to extract a broader range of informative features, and effectively capture the correlation and interaction between complementary modalities. We have performed extensive performance evaluations on various datasets, inpainting algorithms, and post-processing. The results have fully testified the effectiveness, the impressive high generalization ability, and the robustness of our TruVIL scheme compared with the state-of-the-art. While our method demonstrates strong competitiveness in localization performance, there still exist several potential limitations. Our algorithm only analyzes inpainted videos, thus potentially constraining the model's ability to detect uninpainted videos. Additionally, the model's performance may be affected when confronted with discontinuities and rapid dynamic changes in videos. Lastly, our research leverages video data from openly accessible datasets, which feature high quality and stable frame rates, posing challenges in handling complex data sourced from the internet. In future work, we aim to delve deeper into these issues, continuously enhancing the algorithm's robustness and generalization ability to better address real-world needs. Furthermore, considering the current popularity of Large Language Models (LLMs), we intend to explore their potential applications in visual tasks as part of our forward-looking research efforts.

\bibliographystyle{IEEEtran}
\bibliography{lzj_manuscript}
\vfill
\end{document}